\documentclass[runningheads]{llncs}

\usepackage{graphicx}
\usepackage{amsmath} 
\usepackage{amssymb}
\usepackage{physics}
\usepackage[utf8]{inputenc}
\usepackage[english]{babel}
\usepackage{dsfont}

\DeclareMathOperator{\Sup}{\textup{Sup}}

\usepackage{cite}
\usepackage{url}

\makeatletter
\AtBeginDocument{%
  \def\doi#1{\url{https://doi.org/#1}}}
\makeatother

\title{Colour Morphological Distance Ordering based on the Log-Exp-Supremum}

\author{Marvin Kahra\inst{1} and Michael Breu\ss\inst{1}}
 \authorrunning{M. Kahra and M. Breu{\ss}}
 \institute{Institute for Mathematics, Brandenburg Technical University Cottbus-Senftenberg, 03046 Cottbus, Germany \\
 \email{\{marvin.kahra,sridhviv,breuss\}@b-tu.de}}

\begin{document}

\maketitle

\begin{abstract}
    Mathematical morphology, a field within image processing, includes various filters that either highlight, modify, or eliminate certain information in images based on an application's needs. Key operations in these filters are dilation and erosion, which determine the supremum or infimum for each pixel with respect to an order of the tonal values over a subset of the image surrounding the pixel. This subset is formed by a structuring element at the specified pixel, which weighs the tonal values. Unlike grey-scale morphology, where tonal order is clearly defined, colour morphology lacks a definitive total order. As no method fully meets all desired properties for colour, because of this difficulty, some limitations are always present.

    This paper shows how to combine the theory of the log-exp-supremum of colour matrices that employs the Loewner semi-order with a well-known colour distance approach in the form of a pre-ordering. The log-exp-supremum will therefore serve as the reference colour for determining the colour distance.
    To the resulting pre-ordering with respect to these distance values, we add a lexicographic cascade to ensure a total order and a unique result. The objective of this approach is to identify the original colour within the structuring element that most closely resembles a supremum, which fulfils a number of desired properties. Consequently, this approach avoids the false-colour problem.
    The behaviour of the introduced operators is illustrated by application examples of dilation and closing for synthetic and natural images.

\end{abstract}

\section{Introduction} 

Mathematical morphology is a theoretical framework employed for the analysis of spatial patterns in images. Over the years, it has evolved into a highly effective domain of image processing, as detailed in texts such as \cite{Najman-Talbot,Roerdink-2011,Serra-Soille}. Morphological operators are fundamentally composed of two primary elements. The first element is the structuring element (SE), defined by its shape, size, and position, and classified into two categories: flat and non-flat, as referenced in \cite{Haralick_1}. A flat SE specifies a neighbourhood around the central pixel where relevant morphological operations are conducted, whereas a non-flat SE includes an additional mask containing finite values used as additive shifts. Typically, the SE is implemented as a window that moves across the image. The second primary element serves to compare values within an SE. Dilation and erosion are two fundamental operations in mathematical morphology; they assign the maximum or minimum, respectively, within the SE of the discrete image function to the pixel value. Many practical morphological filtering techniques, such as opening, closing, or top hats, are derived through combinations of dilation and erosion. Given that dilation and erosion are duals, algorithm development often relies on just one of these operations.

Dilation and erosion operations play a crucial role in colour morphology, forming the basis for our subsequent analysis. In morphology, a key task is comparing tonal values, which, in our context, refers to comparing colour values within the SE. In simpler areas like greyscale morphology, one can achieve a total order of colour values by directly interacting with complete lattices, as discussed in \cite{Blusseau,Serra-Soille}. However, in colour morphology, there is no natural total ordering for colour values. Consequently, semi-orders and alternative foundational structures are employed, as elaborated in \cite{BAR76}. One potential method is to treat each colour channel of an image independently, applying greyscale morphology individually. A more widely accepted method involves using a vector space model, where colours are represented as vectors within a colour space, referenced in \cite{Angulo}, \cite{APT08}, or \cite{LEZ07}. We will consider the latter method and define a total ordering of the corresponding colours.

We construct our total order according to the model of Angulo \cite{lex_cascade} and introduce a semi-order based on the colour difference to a certain reference colour. This semi-order is then extended to a total order by a corresponding lexicographic cascade. Such approaches have the advantage of avoiding so-called false colours, i.e. colours that were not originally included in the subset in the form of the SE with respect to the pixel under consideration. Angulo used different reference colours in his work depending on the application. In this work, we want to pursue a fixed, more adaptive approach and therefore specify a precise procedure that should determine this reference colour pixel by pixel. We want to use the log-exp-supremum (LES), which was introduced recently in \cite{DGMM24} by Kahra, Breu{\ss}, Kleefeld and Welk. 

The method is based on Burgeth and Kleefeld's method \cite{Loewner}, which used the bi-cone shape of a modified HCL colour space in conjunction with the Loewner semi-order to identify colours as symmetrical matrices and cones in the bi-cone. A colour was considered to have a higher value if its cone included the cone of the smaller colour. Since it was a semi-order, the approximation of the maximum by means of Maslov's log-sum-exp representation \cite{Maslov} was used in \cite{DGMM24} and \cite{LES-Journal} to select from the set of upper bounds one that is minimal with respect to an auxiliary order function.

In this paper we want to combine the procedures of \cite{lex_cascade} and \cite{LES-Journal} in order to obtain a new distance-based ordering scheme for colour morphology. Since the LES operates in the modified HCL and this is closely related to the HSL colour space, which is used to a large extent in \cite{lex_cascade}, this allows us a relatively direct transfer of this methodology. The use of the LES is particularly suitable here because, as was shown in \cite{DGMM24}, it leads to a brightening of the colours and experience has shown that the luminance in the HCL/HSL plays the most important role with regard to the order relation for the individual components. As the distance function used plays an important role in this method, we will compare three different distance functions with each other and also compare the method with more classic methods in RGB colour space, both distance-based and channel-wise.

\section{Colour Morphology}

In order to make this paper self-contained, we briefly recall some basics of mathematical morphology in general and colour morphology in particular. Then we will use the transformation of the paper \cite{Loewner} to transform the colour vectors into symmetric matrices. These will be the starting point and together with the Loewner semi-order as the main means to determine the supremum, as described at the end of this section, or reference colour later on.

For the greyscale case, we begin with a two-dimensional discrete image domain $\Omega \subseteq \mathbb{Z}^2$ and a single channel greyscale image given by a function $f: \Omega \rightarrow [0,255]$ that assigns a discrete greyscale value to every pixel in the image domain $\Omega$. In general, we can assume to have a non-flat morphology, i.e. the SE can be represented as a function $g: \mathbb{Z}^2 \rightarrow \mathbb{R} \cup \{ - \infty\}$ with 
\begin{align}
    g(\boldsymbol x) := 
    \begin{cases}
        \eta(\boldsymbol x), \quad & \boldsymbol x \in G_0, \\
		- \infty, & \text{otherwise},
    \end{cases}
    \quad G_0 \subset \mathbb{Z}^2, \quad \boldsymbol{x} \in \mathbb{Z}^2,
    \label{SE}
\end{align}
where $\eta: \mathbb{Z}^2 \rightarrow \mathbb{R}$ assigns an offset and $G_0$ is a set centred at the origin. Whereas in the special case of flat morphology, we have $\eta(\boldsymbol x) = 0$ for all $\boldsymbol{x} \in \mathbb{Z}^2$. To explain a filter in mathematical morphology, one can combine the image function with one or various SEs by one or more operations. Two of the most elementary operations for this are dilation 
\begin{align}
    (f\oplus g)(\boldsymbol x) := \max_{\boldsymbol u\in \mathbb{Z}^2}  {\{f(\boldsymbol x- \boldsymbol u)\: +\: g(\boldsymbol u) \}}, \quad \boldsymbol{x} \in \Omega,
    \label{dilation}
\end{align}
and erosion
\begin{align}
    (f\ominus g)(\boldsymbol x) := \min_{\boldsymbol u\in \mathbb{Z}^2}  {\{f(\boldsymbol x + \boldsymbol u)\: -\: g(\boldsymbol u) \}}, \quad \boldsymbol{x} \in \Omega.
    \label{erosion}
\end{align}
Note that these operations are generally explained using supremum and infimum, but in the discrete case shown here these correspond to the maximum and minimum. In particular, these two operations are dual to each other in this case. Let the negative image $\overline{f}$ of $f$ be explained as
\begin{align}
    \overline{f}(\boldsymbol{x}) = f_{\max} - f(\boldsymbol{x}) + f_{\min}  \quad \forall \boldsymbol{x} \in \Omega,
    \label{neg_img}
\end{align}
where $f_{\max}, f_{\min}$ are the largest and the smallest value of the image $f$.
Then the duality of the two applies according to
\begin{align}
    (\overline{f} \ominus g)(\boldsymbol{x}) = \overline{(f \oplus g)}(\boldsymbol{x}) \quad \text{or} \quad (f \ominus g)(\boldsymbol{x}) = \overline{(\overline{f} \oplus g)}(\boldsymbol{x}) \quad \forall \boldsymbol{x} \in \Omega. 
    \label{duality}
\end{align}

The colour morphology resembles the greyscale morphology but with the difference that we no longer have just one channel, but three. For an overview of some representation formats see \cite{SHA03}. The simplest approach is a channel-wise processing of the image with the RGB colour model or any other colour space that is based on the same underlying colour model like HSL, see e.g. \cite{Srid} for a recent example of a channel-wise RGB scheme implementation. Since this approach is the same as in the greyscale case, just in three dimensions, it maintains the duality of dilation and erosion. 

Based on the idea of Burgeth and Kleefeld, see \cite{Loewner}, we use a modification of the HSL or RGB colour space, which uses a modified luminance and the chroma variable. This provides a better geometric illustration of the Loewner semi-order used in this colour space. As starting point for the transformation to this colour space, we assume the normalised RGB space $[0, 1]^3 = [0, 1] \times [0, 1] \times [0, 1]$. Then, we transform the RGB space to the general HCL space, according to
\begin{align*}
    &H := 
    \begin{cases}
        \frac{G - B}{6C} \mod 1, & \textup{ if } M = R, \\
        \frac{B - R}{6C} + \frac{1}{3} \mod 1, & \textup{ if } M = G, \\
        \frac{R - G}{6C} + \frac{2}{3} \mod 1, & \textup{ if } M = B.
    \end{cases}
    , \quad M := \max\{R,G,B\}, \\
    &C := M - m, \quad m := \min\{R,G,B\} \text{ and}\\
    &L := \frac{1}{2} (M + m).
\end{align*}
The luminance $L$ will be replaced by the modified luminance $L_m := 2L-1 = M + m - 1$. Since these are cylindrical coordinates and the colour space a bi-cone with vertices in $(0, 0, 1)$ and $(0, 0, -1)$, we can also transform the coordinates into the corresponding Cartesian coordinates through $x = C \cos(2\pi H)$, $y = C \sin(2\pi H)$ and $z = L_m$.

However, if we consider every channel separately from each other, we lose the correlated information between the channels. To circumvent this behaviour, we convert these vectors into symmetric matrices and vice versa using 
\begin{align}
    \begin{pmatrix}
        x \\
        y \\
        z
    \end{pmatrix}
    \longleftrightarrow \frac{1}{\sqrt{2}}
    \begin{pmatrix}
        z - y \hspace{0.2cm}& x \\
        x \hspace{0.2cm}& z + y
    \end{pmatrix}
    \quad \text{or} \quad 
    \begin{pmatrix}
        \alpha \hspace{0.2cm}& \beta \\
        \beta \hspace{0.2cm} & \gamma
    \end{pmatrix}
    \longleftrightarrow \frac{1}{\sqrt{2}}
    \begin{pmatrix}
        2\beta \\
        \gamma - \alpha \\
        \gamma + \alpha
    \end{pmatrix}.
    \label{matrix_transformation}
\end{align}

\noindent Since neither higher-dimensional vectors nor matrices have a clearly defined order relation, it is necessary to use a semi-order. In the case of the symmetric matrices considered here, the Loewner semi-order is suitable, see \cite{lowner_monotone}, which is given by
\begin{align}
    \boldsymbol{A} \geq_\textup L \boldsymbol{B} :\Longleftrightarrow \boldsymbol{A} - \boldsymbol{B} \in \textup{Sym}_+(n), \quad \boldsymbol{A}, \boldsymbol{B} \in \textup{Sym}(n),
        \label{loewner_order}
\end{align}
where $\textup{Sym}(n)$ is the set of symmetric matrices of $\mathbb{R}^{n \times n}$, $n \in \mathbb{N}$, and $\textup{Sym}_+(n)$ the corresponding set of positive semi-definite symmetric matrices. This order arises naturally due to the fact that $\textup{Sym}(n)$ represents a convex cone, see \cite{convex_Analysis}, that induces this partial order. Since this is a semi-order and therefore only provides a set 
\begin{align*}
    \mathcal{U}(\mathcal{X}) := \left\{ \boldsymbol Y \in \text{Sym}(2) : \boldsymbol X \leq_\textup L \boldsymbol Y \quad \forall \boldsymbol X \in \mathcal{X} \subset \text{Sym}(2) \right\},
\end{align*}
of upper bounds instead of a supremum, it is usually necessary to define a minimiser to form a supremum. However, due to the direct supremum calculation from \cite{LES-Journal}, which we summarise in the next section, we will not need this. Instead, the calculated supremum for the case of non-negative eigenvalues will be a minimiser with respect to the two-dimensional space of eigenvalues of our symmetric matrices $\mathcal{X}$ equipped with the lexicographic order for the intersection 
\begin{align*}
    \mathcal{U}_*(\mathcal{X}) := \bigcap_{p > 0} \mathcal{U}_p(\mathcal{X}) := \bigcap_{p > 0} \{ \boldsymbol{Y} \in \textup{Sym}(2) : \boldsymbol Y^p \in \mathcal{U}(\mathcal{X}^p) \}
\end{align*}
of all exponentiated upper bounds of the matrices. The other case for negative eigenvalues follows by a simple translation, see \cite{LES-Journal}. For a more detailed study of the other possible minimisers mentioned, we refer to \cite{WelkQuantile}.

\section{Procedure of the Distance based Log-Exp-Supremum}

In this section, we want to present the individual steps of the procedure and also point out alternatives. We will briefly summarise the most relevant steps of the procedure from \cite{LES-Journal} and link them to our approach. We shall then introduce the distance that we intend to use and the lexicographic cascade.     

\subsection{The log-exp-supremum}

The construction of the LES is based on the well known log-sum-exp-approximation of the maximum 
\begin{align}
    \max\{x_1, \dots, x_n\} = \lim_{m \rightarrow \infty} \frac{1}{m} \log \left( \sum_{i = 1}^n \mathrm{e}^{m x_i} \right), \quad x_1, \dots, x_n \in \mathbb{R},
    \label{LogSumExp}
\end{align}
which was introduced by Maslov in \cite{Maslov} for convex optimisation. In this case, it needs to be formulated for matrices. Therefore, we will use the multi-set $\mathcal{X} = \{\boldsymbol X_1,\dots , \boldsymbol X_n\}$ as our input data for the dilation and define the LES as
\begin{align}
    \boldsymbol S := \Sup_{\textup{LE}}(\mathcal{X}) := \lim_{m \rightarrow \infty} \frac{1}{m} \log \left( \sum_{i=1}^n \mathrm{e}^{m \boldsymbol X_i} \right).
    \label{LES}
\end{align}
For the direct computation of this supremum, we need the spectral decomposition of the matrices of $\mathcal{X}$: 
\begin{align}
    \begin{split}
        &\boldsymbol X_i = \lambda_i \boldsymbol u_i \boldsymbol u_i^\textup{T} + \mu_i \boldsymbol v_i \boldsymbol v_i^\textup{T}, \quad \lambda_i \geq \mu_i, \quad \langle \boldsymbol u_i, \boldsymbol v_i \rangle = 0, \quad \abs{\boldsymbol u_i} = 1 = \abs{\boldsymbol v_i}, \\
        &\boldsymbol u_i = (c_i,s_i)^\textup{T} , \quad \boldsymbol v_i = (-s_i,c_i)^\textup{T}, \quad  c_i = \cos(\varphi_i), \quad s_i = \sin(\varphi_i),\\
        &\varphi_i \in \left[- \frac{\pi}{2}, \frac{\pi}{2}\right], \quad i = 1,\dots,n,
    \end{split}
    \label{spectral_decomposition}
\end{align}
where $\lambda_i,\mu_i \in \mathbb{R}$ are the eigenvalues of $\boldsymbol X_i$ and $\boldsymbol u_i \perp \boldsymbol v_i$ are the unit eigenvectors to the corresponding eigenvalues. We summarise all eigenvectors of this spectral decomposition for the matrices of $\mathcal{X}$ to $\mathcal{V}$ and declare the set $\mathcal{V}_{\sup}$ of all eigenvectors with a corresponding eigenvalue that is one of the largest eigenvalues of $\mathcal{X}$ as $\mathcal{V}_{\sup} := \left\{ \boldsymbol{v} \in \mathcal{V} : \lambda_{\boldsymbol{v}} \geq \lambda_i, ~i \in \{1, \dots, n\} \right\}$, where $\lambda_{\boldsymbol{v}}$ is the corresponding eigenvalue of $\boldsymbol{v}$.
According to \cite{LES-Journal}, this then results in
\begin{align}
    \boldsymbol{S} = 
    \begin{cases}
        \lambda_1 \boldsymbol{I} &, \text{ if } \lambda_1 \text{ is not unique and }  \exists \boldsymbol{v} \in \mathcal{V}_{\sup}: ~\boldsymbol v \neq \pm\boldsymbol u_1, \\
        \lambda_1 \boldsymbol{u}_1 \boldsymbol{u}_1^{\textup T} + \mu_* \boldsymbol{v}_1 \boldsymbol{v}_1^{\textup T} &, \text{ otherwise},
    \end{cases}
    \label{LES_calc}
\end{align}
where $\lambda_1$ is (one of) the largest eigenvalues of $\mathcal{X}$, $\mu_* \leq \lambda_1$ is the next largest eigenvalue of $\mathcal{X}$ whose eigenvector $\boldsymbol{v}_* \neq \pm \boldsymbol{u}_1$ and $\boldsymbol{I}$ is the identity matrix. 

This supremum formation allows the formation of a corresponding infimum by means of a duality argument as in \eqref{duality}. Furthermore, it is also transitive and as such it fulfils the associativity of dilation and the relation 
\begin{align*}
    (\boldsymbol{f} \ominus \boldsymbol{g}_1) \ominus \boldsymbol{g}_2  = \boldsymbol{f} \ominus (\boldsymbol{g}_1 \oplus \boldsymbol{g}_2)
\end{align*}
of erosion, see Proposition 1 and Theorem 3 in \cite{LES-Journal}. The property of transitivity is typically only observed in greyscale morphology or channel-based colour morphology derived from it. Consequently, this supremum represents a rare opportunity to transfer the transitivity to colour morphology in a non-channel-based application. By employing symmetric matrices in contrast to the channel-wise approach, all channels are considered concurrently, and no correlated information between the channels is lost.

\subsection{The distance based ordering}

Now that we have introduced a supremum, which gives us the ``actual" target colour, we can use this as a reference for the available colours in the current environment of the SE. For our approach, this means that we want to select the colour within the SE that is most similar to the reference colour. In order to quantify this similarity, we introduce a distance term and, since we are dealing with colours, the colour difference $\Delta E$ is suitable for this purpose. 

As ordering by a distance function is a common approach, there are many different possible definitions of colour difference, most of which are based on the underlying colour space. The simplest and most common is to use the $\textup L_1$ or $\textup L_2$ distance. However, the problem with these is that all components are considered equally, which is not the case with dilation, for example.  This is because the brightness of a colour should have a higher weight than the hue. 

We therefore first consider a modification of the HyAB distance, which is used in L$^*$a$^*$b$^*$ colour spaces and demonstrably delivers better results there than the $\textup L_2$ distance, see \cite{Lab_dist}. Since the a$^*$ and b$^*$ components can be translated directly to the $x$ and $y$ of section 2, we will just replace the luminance with the modified luminance and obtain the modified HyAB distance as follows: 
\begin{align}
\begin{split}
    &\Delta E_{\textup{mHyAB}} \big(\boldsymbol{f}_1, \boldsymbol{f}_2\big) := \sqrt{(x_1 - x_2)^2 + (y_1 - y_2)^2} + \abs{z_1 - z_2}, \\ 
    &\boldsymbol{f}_i = (x_i, y_i, z_i), \quad i = 1,2,
    \label{HyAB_dist}
\end{split}
\end{align}
where we used the Cartesian representation for colours according to Section 2. Since the LES, as can already be seen in \cite{DGMM24}, can deviate greatly from the colours used, which may lead to a fading towards the white spectrum of the colours, the colour differences $\norm{\boldsymbol{f}_1 - \boldsymbol{f}_2}$ can be quite high. This is another reason to use the HyAB or mHyAB distance term, as it provides significantly better results \cite{Lab_dist} for larger distances $\norm{\boldsymbol{f}_1 - \boldsymbol{f}_2}_2 \geq 10$ than the use of the $\textup L_2$ distance, at least in L$^*$a$^*$b$^*$.

To use the modified HCL directly for this, we need to evaluate the hue component. As this is an angle, it does not allow any direct ordering. However, we can instead look at the angular distance \cite{angular_morphology}, which is given for any two angles $h_1, h_2 \in [0, 2\pi)$ as
\begin{align}
    h_1 \divisionsymbol h_2 := 
    \begin{cases}
        \abs{h_1 - h_2}, &\text{ if } \abs{h_1 - h_2} \leq \pi,\\
        2\pi - \abs{h_1 - h_2}, &\text{ otherwise}.
    \end{cases}
    \label{angular_dist}
\end{align}
We use the hue component $H^{\boldsymbol{S}}$ of the LES $\boldsymbol{S}$ according to $\boldsymbol S \longleftrightarrow \boldsymbol{f}_{\boldsymbol{S}} = (H^{\boldsymbol{S}}, C^{\boldsymbol{S}}, L_m^{\boldsymbol{S}})$ as reference for the ordering of the hue components by computing $H \divisionsymbol H^S$.
However, with regard to the HSL colour space, we encounter the problem of the instability of the hue component at low saturation, which is solved by scaling with the average saturation, as proposed in \cite{lex_cascade}. Due to the close relationship between saturation and chroma, we replace the saturation here and obtain one of the simplest distances in this respect:
\begin{align}
    \begin{split}
        &\Delta E_{1H} \big( \boldsymbol{f}_1, \boldsymbol{f}_2 \big) := \abs{L_{m_1} - L_{m_2}} + \abs{C_1 - C_2} + \frac{\abs{C_1 - C_2}}{2} \abs{H_1 \divisionsymbol H_2}, \\
        &\boldsymbol{f}_i = (H_i, C_i, L_{m_i}), \quad i = 1,2.
    \end{split}
    \label{simple_hc_dist}
\end{align}
This distance function represents a variation of the $\textup L_1$ distance weighted for the hue component with the chroma component. Another possibility is the polar distance 
\begin{align}
    \begin{split}
        &\Delta E_{\textup{polar}} \big( \boldsymbol{f}_1, \boldsymbol{f}_2 \big) := \sqrt{(L_{m_1} - L_{m_2})^2 + C_1^2 + C_2^2 - 2C_1C_2 \cos(H_1 \divisionsymbol H_2)}, \\
        &\boldsymbol{f}_i = (H_i, C_i, L_{m_i}), \quad i = 1,2,
    \end{split}
    \label{polar_dist}
\end{align} 
also proposed in \cite{lex_cascade} for the HSL. This is the $\textup L_2$ distance in cylindrical coordinates and using the notation for the angular distance. 

By calculating the reference colour in the form of the LES \eqref{LES_calc} and one of the distance functions \eqref{HyAB_dist}, \eqref{simple_hc_dist} or \eqref{polar_dist} with regard to the colour difference to this reference colour, we have declared a pre-order. However, in certain cases, such as with higher iterations of the morphological operators, two colours in the SE may have the same distance. We therefore need at least one more criterion to define a total order. 

To resolve this issue, we use a simple lexicographic cascade. Experience with the HSL colour space has shown that the cascade $L \rightarrow S \rightarrow -(h \divisionsymbol h_0)$ is preferable, where $h = 2\pi H \in [0, 2\pi)$ is the angular of the colour and $h_0$ is the angular of the reference colour, see \cite{lex_cascade}. Due to the close relationship between HCL and HSL, we also adopt the corresponding cascade here and simply exchange the saturation with the chroma. Finally, we obtain the following total order relation for a distance based log-exp-supremum (DLES):
\begin{align}
\begin{split}
    &\boldsymbol{f}_1 <_{\textup{DLES}} \boldsymbol{f}_2 \Longleftrightarrow
    \begin{cases}
        \Delta E \big(\boldsymbol{f}_{\textup{LES}}, \boldsymbol{f}_1\big) < \Delta E \big(\boldsymbol{f}_{\textup{LES}}, \boldsymbol{f}_2\big) \text{ or } \\
        \Delta E \big(\boldsymbol{f}_{\textup{LES}}, \boldsymbol{f}_1\big) = \Delta E \big(\boldsymbol{f}_{\textup{LES}}, \boldsymbol{f}_2\big) \text{ and } \\
        \begin{cases}
            L_{m_1} < L_{m_2} \text{ or } \\
            L_{m_1} = L_{m_2} \text{ and } C_1 < C_2 \text{ or } \\
            L_{m_1} = L_{m_2} \text{ and } C_1 = C_2 \text{ and } H_1 \prec H_2,
        \end{cases}
    \end{cases} 
    \\
    &\boldsymbol{f}_i = \boldsymbol{f}(\boldsymbol{x}_i) = (H_i, C_i, L_{m_i})^{\textup T}, \quad \boldsymbol{x}_i \in \Omega, \quad i = 1,2, 
\end{split}
\label{DLES_order}
\end{align}
where $\boldsymbol{f}_{\textup{LES}}$ is the corresponding colour to the calculated LES $\boldsymbol{S}$ from \eqref{LES_calc} and $\Delta E$ represents one of the three distance functions \eqref{HyAB_dist}, \eqref{simple_hc_dist} or \eqref{polar_dist} and a possible transformation of the colours in the case of \eqref{HyAB_dist}. Furthermore, the relation used for the hue component is according to \cite{lex_cascade} explained as
\begin{align*}
    \begin{split}
        &H_1 \prec H_2 \Longleftrightarrow
        \begin{cases}
            -(h_1 \divisionsymbol h_{\textup{LES}}) < -(h_2 \divisionsymbol h_{\textup{LES}}) \text{ or}\\
            -(h_1 \divisionsymbol h_{\textup{LES}}) = -(h_2 \divisionsymbol h_{\textup{LES}}) \text{ and } (h_1 - h_{\textup{LES}}) \textup{ mod } 2\pi \leq \pi,
        \end{cases}\\
        &h_i = 2\pi H_i, \quad i = 1,2, \quad h_{\textup{LES}} = 2\pi H_{\textup{LES}}.
    \end{split}
\end{align*}
This additional priority to the angular distance is necessary for a total order, as the angular distance for two permissible angles is the same if they differ by $\frac{\pi}{2}$ and $\frac{3\pi}{2}$ of $h_{\textup{LES}} $.

\section{Experiments}

In this section, we will compare the presented DLES with a similar conventional method, namely a channel-wise distance-based supremum formation in RGB, and compare the specified distance functions for the DLES. To do this, we will use both synthetic and natural colour images, which have different dimensions and apply flat square SEs of different sizes, which are centred at the origin, to them.

\subsection{Comparison with a channel-wise approach}

\begin{figure}[t]
\centering
\minipage{0.3\linewidth}
 \includegraphics[width=\linewidth]{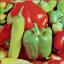}
\endminipage
\hfill
\minipage{0.3\linewidth}
 \includegraphics[width=\linewidth]{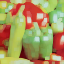}
\endminipage
\hfill
\minipage{0.3\linewidth}
 \includegraphics[width=\linewidth]{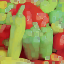}
\endminipage
\hfill
\\[0.6em]
\minipage{0.3\linewidth}
 \includegraphics[width=\linewidth]{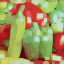}
\endminipage
\hfill
\minipage{0.3\linewidth}
 \includegraphics[width=\linewidth]{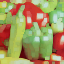}
\endminipage
\hfill
\minipage{0.3\linewidth}
 \includegraphics[width=\linewidth]{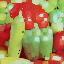}    
\endminipage
\caption{\label{dilation_comparison} Comparison of dilation results with channel-wise RGB approaches and the DLES method for different distance functions of a $64 \times 64$ pepper image with a $3 \times 3$ square SE. 
{\bf Top:} {\bf From left to right:} Original image, channel-wise Matlab dilation and distance based dilation with the reference colour white in RGB space.
{\bf Bottom:} {\bf From left to right:} Dilation with the DLES method using the distance function $\Delta E_{\textup{mHyAB}}$, $\Delta E_{\textup{polar}}$ and $\Delta E_{1H}$.
}
\end{figure}

For this experiment, we use the pepper image downscaled to $64\times64$ pixels and a $3\times3$ SE to perform one dilation at a time. The DLES dilations are based on the total order \eqref{DLES_order} with the corresponding distance functions. In the case of the channel-wise approach, we use the classic $\textup{L}_2$ distance and a cascading order according to \cite{lex_cascade} with white as the reference colour. In addition, we compare them with the standard channel-wise dilation of Matlab in RGB, see Figure \ref{dilation_comparison}. 

We can see here that the distance-based channel-wise dilation with the $\textup L_2$ distance and the DLES variant with $\Delta E_{1H}$ in particular do not deliver satisfactory results. With the first of the two, almost all bright colours in the image are replaced by more saturated colours. This contradicts our expectation of dilation, which should expand the more white pixel sets in particular. The latter succeeds in retaining the bright areas to some extent, but we see clear fragments appearing in unexpected places. Furthermore, we also see an irregular behaviour here, for example in the long green pepper stem, which becomes thicker, which indicates that the green colours are rated higher than the red ones, but at the same time it also becomes shorter, which suggests the opposite.

The two other DLES images, based on $\Delta E_{\textup{mHyAB}}$ and $\Delta E_{\textup{polar}}$, look very similar apart from a few minor details. Otherwise, both show the expected behaviour of a dilation with a noticeable expansion of the bright areas, as with the Matlab dilation. However, in contrast to the latter, they do not exhibit any false colours, which appear particularly at the edge of green to red areas and tend to become yellowish. Due to this similarity between the two DLES variants with $\Delta E_{\textup{mHyAB}}$ and $\Delta E_{\textup{polar}}$, we will compare them more closely in the next experiment.

\subsection{Comparison of the DLES variants}

\begin{figure}[t]
\centering
\minipage{0.2\linewidth}
 \includegraphics[width=\linewidth]{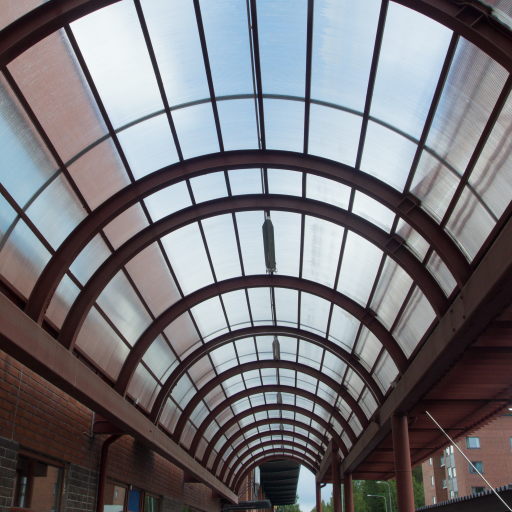}
\endminipage
\hspace{0.5cm}
\minipage{0.2\linewidth}
 \includegraphics[width=\linewidth]{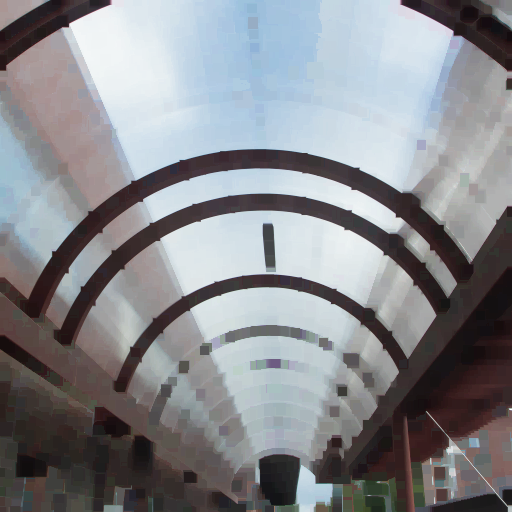}
\endminipage
\hspace{0.5cm}
\minipage{0.2\linewidth}
 \includegraphics[width=\linewidth]{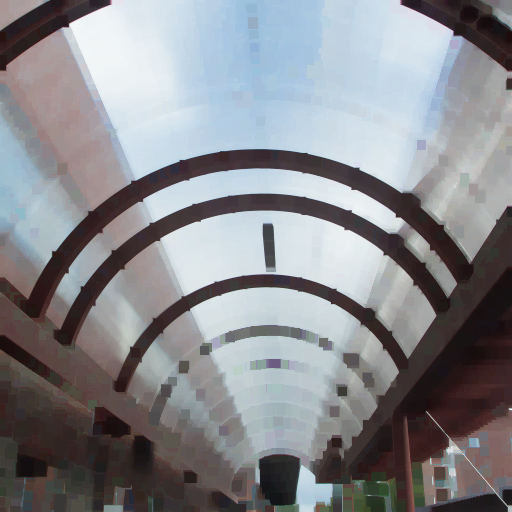}
\endminipage
\hspace{0.5cm}
\minipage{0.2\linewidth}
 \includegraphics[width=\linewidth]{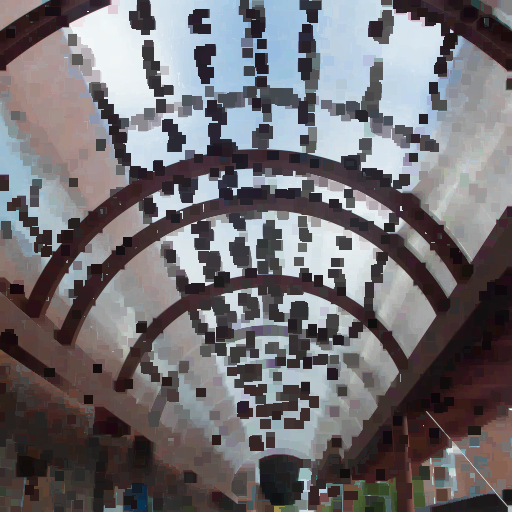}
\endminipage
\caption{\label{closing_comparison} Comparison of closing results for a $512 \times 512$ image from the TAMPERE17 noise-free image database \cite{imageDB} with a $9 \times 9$ SE.
{\bf From left to right:} Original image, closing with the DLES method using the functions $\Delta E_{\textup{mHyAB}}$, $\Delta E_{\textup{polar}}$ and $\Delta E_{1H}$. 
}
\end{figure}

In this experiment we will look at the behaviour of the DLES variants during the closing operation. This is generally applied by a dilation followed by an erosion. The erosion is obtained by the duality of the LES, i.e. by choosing the colour that is closer to the infimum or the dual supremum. The lexicographic cascade can then also be determined analogously using the duality. For this, we use an image with the dimensions $512\times512$ and a $9\times9$ square SE, see Figure \ref{closing_comparison}. 

With a closing, one may expect thinner and darker areas to be closed by lighter ones. Consequently, in our test image, the thinner frames of the glass roof in particular are supposed to vanish. Here, too, we find that the $\Delta E_{\textup{mHyAB}}$ and $\Delta E_{\textup{polar}}$ approaches produce the desired result and even appear visually similar to each other. However, this similarity could also be due to the higher resolution image, as the deviation of individual pixels is less significant. Furthermore, the $\Delta E_{1H}$ approach again delivers a poor result, showing many fluctuations in the actually straight frames.  

\begin{figure}[t]
\centering
\minipage{0.32\linewidth}
 \includegraphics[width=\linewidth]{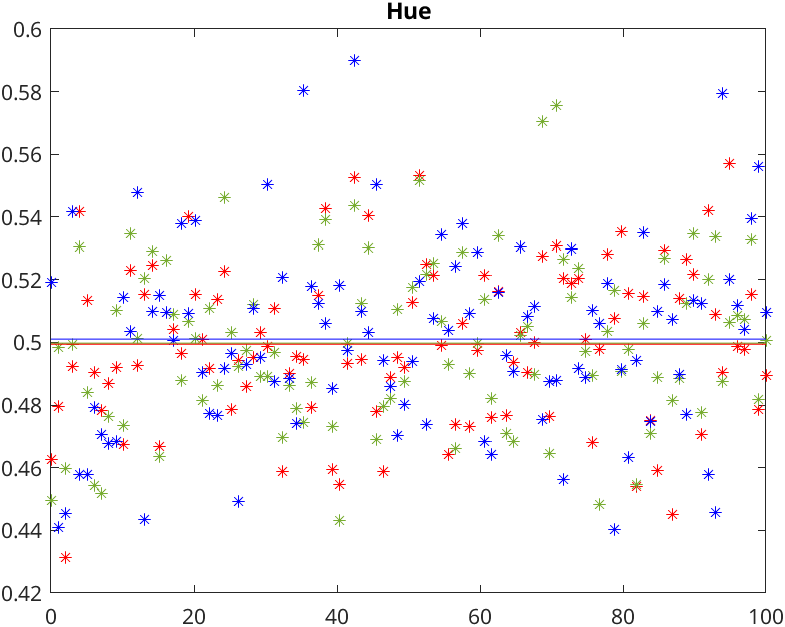}
\endminipage
\hfill
\minipage{0.32\linewidth}
 \includegraphics[width=\linewidth]{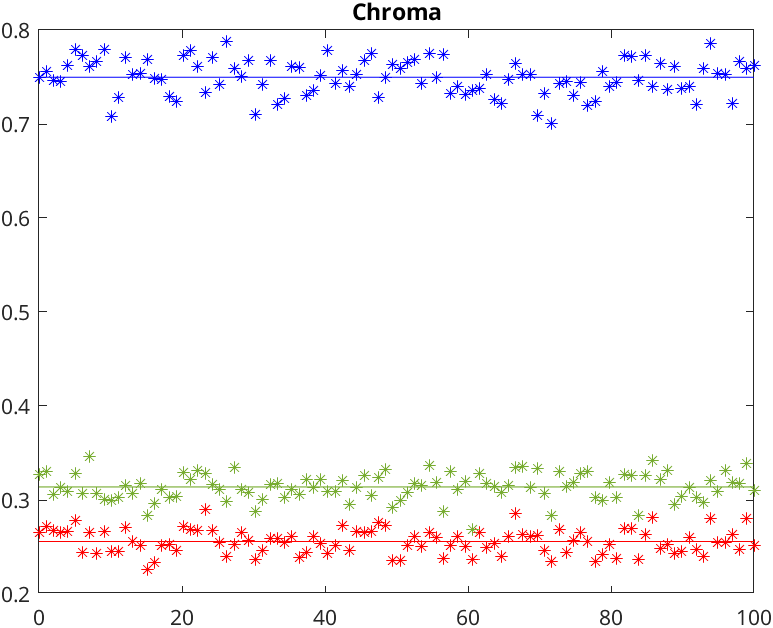}
\endminipage
\hfill
\minipage{0.32\linewidth}
 \includegraphics[width=\linewidth]{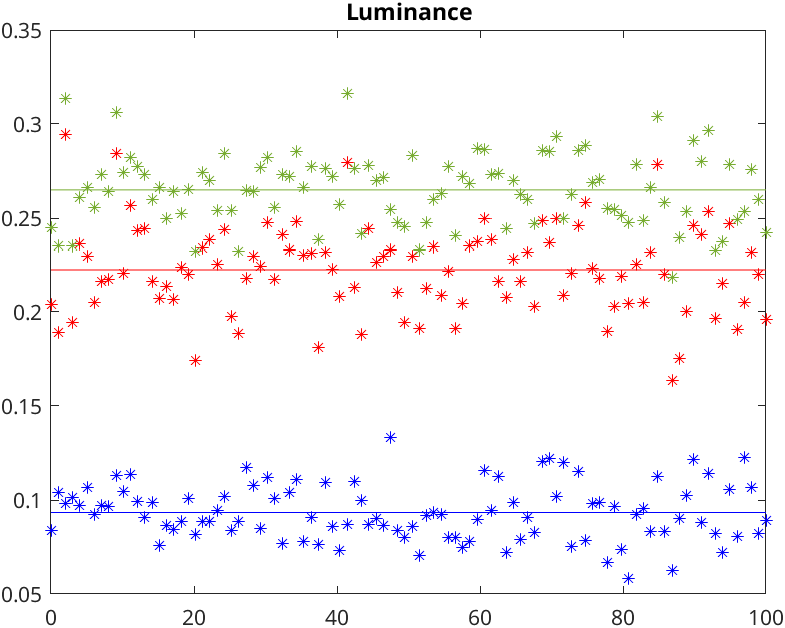}
\endminipage
\hfill
\caption{\label{rand_closing_comparison} Component-wise comparison of closing results for one hundred randomly generated $32 \times 32$ images with the DLES method using $\Delta E_{\textup{mHyAB}}$ (red), $\Delta E_{\textup{polar}}$ (green) and $\Delta E_{1H}$ (blue) and a $3\times 3$ SE.
{\bf From left to right:} Hue, chroma and modified luminance with the corresponding mean values over all images. 
}
\end{figure}

In order to shed more light on the differences between the DLES variants, we will examine how the individual components behave on average. To do this, we perform the closing operation on one hundred randomly generated $32\times32$ pixel images, see Figure \ref{rand_closing_comparison}.

Here we can generally see that the distances $\Delta E_{\textup{mHyAB}}$ and $\Delta E_{\textup{polar}}$ actually seem to behave very similarly and show a big difference to the distance $\Delta E_{1H}$. We can conclude that the use of the $\Delta E_{\textup{polar}}$ leads to brighter and more saturated colours compared to the use of $\Delta E_{\textup{mHyAB}}$. However, these differences are generally quite small. As far as the course of the mean values is concerned, it can be seen, especially with the modified luminance, that the courses are similar, but the polar distance $\Delta E_{\textup{polar}}$ appears to show more continuous mean changes than the distance $\Delta E_{\textup{mHyAB}}$, whose points show a somewhat greater jumping behaviour.

With this method, however, it should be noted that all operations are approximations of these, which have a certain degree of error. This can be seen, for example, in the closing operation, which should not actually change when used repeatedly with the same SE. However, if we carry out a second closing after the first and calculate the total difference between these two images, there is a statistical deviation of about $0.36\%$, $0.58\%$ and $1.21\%$ per pixel, respectively, for the corresponding distances $\Delta E_{\textup{mHyAB}}$, $\Delta E_{\textup{polar}}$ and $\Delta E_{1H}$. This shows that although this property of the closing is not fulfilled, it at least comes very close.

\section{Conclusion and Future Work}

In this work, we have used the distance based total ordering by Angulo \cite{lex_cascade} with the resulting colour of the LES of \cite{LES-Journal} as reference colour. In contrast to Angulo, we used the modified HCL colour space instead of the HSL colour space and compared three different distance functions for this. Our experiments showed that two of these three distance functions produced good results, which are similar to those of a classical channel-wise approach, but do not produce false colours. The resulting morphological operations do not fulfil all the structural properties of the exact algebraic morphological operations, but the difference seems negligible. 

In the future, the question of whether there are even better distance functions that could be used depending on the area of application can be addressed. Alternatively, the distance functions presented here also offer scope for more detailed investigation, for example to determine the slightly different behaviour of the two distance functions $\Delta E_{\textup{mHyAB}}$ and $\Delta E_{\textup{polar}}$ more precisely and thus to be able to make a better prediction as to which one should be used in an application. Furthermore, the difference mentioned at the end of the second experiment raises the question of how this error changes for other or concatenated operations.

\bibliographystyle{splncs04}
\bibliography{MyBibliography}

\end{document}